\documentclass{article}

\usepackage[preprint]{icml2026}
\usepackage{microtype}
\usepackage{booktabs}
\usepackage{graphicx}
\usepackage{amsmath,amssymb,amsthm}
\newtheorem{proposition}{Proposition}
\usepackage[hidelinks]{hyperref}
\hypersetup{pdfsubject={Benchmark reliability under model quantization}}

\icmltitlerunning{A Reliability Audit for 4-bit Quantization Benchmarks}

\begin{document}

\twocolumn[
\icmltitle{Pre-Registering the Detectable Effect:\\A Paired-MDE Budget for 4-bit Quantization Benchmarks, with a Pilot Audit}
\hypersetup{pdftitle={Pre-Registering the Detectable Effect: A Paired-MDE Budget for 4-bit Quantization Benchmarks, with a Pilot Audit}}

\icmlsetsymbol{equal}{*}

\begin{icmlauthorlist}
\icmlauthor{Zexin Zhuang}{smu}
\icmlauthor{Yanhang Li}{neu}
\icmlauthor{Zhichao Fan}{uiuc}
\end{icmlauthorlist}

\icmlaffiliation{smu}{Southern Methodist University}
\icmlaffiliation{neu}{Northeastern University}
\icmlaffiliation{uiuc}{University of Illinois Urbana-Champaign}

\icmlcorrespondingauthor{Zexin Zhuang}{zexinz@smu.edu}

\icmlkeywords{quantization, benchmark reliability, minimum detectable effect, prompt sensitivity, evaluation methodology, language models}

\vskip 0.3in
]

\printAffiliationsAndNotice{}

\begin{abstract}
This is a planning-method note with an unpaired pilot audit. We adapt the classical paired-binary sample-size calculation \citep{miettinen1968matched} to quantization benchmarks, giving a conservative \emph{minimum detectable effect} (MDE) bound $\delta^{*} \le (z_{1-\alpha/2}+z_{1-\beta})\sqrt{\rho_d/m}$ in the paired item count $m$ and the FP16$\leftrightarrow$NF4 disagreement rate $\rho_d$. The bound turns ``how reliable is my quantization claim?'' into a one-line budget a benchmark designer can commit to \emph{before running}. We illustrate the bound on four models and four benchmarks ($k{=}5$ splits of $n{=}100$), and add a parallel MMLU prompt-template study to put the bound's quantization-noise scale alongside the prompt-noise scale. Assuming $\rho_d{=}0.10$ (an unmeasured planning value), all observed NF4$-$FP16 deltas fall below the implied MDE, and most cross-split SDs lie within $\pm 1.5$\,pp of the binomial reference $\sqrt{p(1-p)/n}$, so much of the variance reported as ``benchmark unreliability'' on $n{=}100$ subsamples is binomial sampling noise. The single borderline cell (OPT--WinoGrande, $|\Delta|{=}3.2$\,pp) is below the implied MDE at $\rho_d{=}0.10$ but above it at $\rho_d{=}0.05$, illustrating the planning trade-off the bound makes explicit. \textbf{On MMLU, prompt-template ranges of 2--10\,pp meet or exceed the largest observed quantization delta (3.2\,pp), so a quantization audit that does not first fix the prompt template absorbs template variance into its noise floor.} We complement the bound with a five-line pre-registration template.\end{abstract}

\section{Introduction}

Post-training quantization is now the default deployment path for large language models on consumer hardware. NF4 quantization \citep{dettmers2024qlora} packs each weight into one of 16 levels chosen to be optimal for normally distributed weights; GPTQ \citep{frantar2023gptq}, AWQ \citep{lin2023awq}, SmoothQuant \citep{xiao2023smoothquant}, SpQR \citep{dettmers2023spqr}, and ZeroQuant \citep{yao2022zeroquant} have all reported small-to-modest accuracy losses on standard benchmarks. A typical quantization study reports a single accuracy number per model$\times$benchmark cell and concludes that 4-bit inference ``preserves'' or ``slightly degrades'' performance.

This practice elides a basic statistical question: what is the smallest quantization effect that the evaluation protocol could reliably detect in the first place? With $n$ benchmark items and a baseline accuracy of $p$, even a perfectly faithful estimator has standard error at least $\sqrt{p(1-p)/n}$. With $k$ non-overlapping splits, the cross-split standard deviation has expected scale $\sqrt{p(1-p)/n}$ unless the benchmark itself is structured (e.g., MMLU's 57 subjects make subject mix vary across random splits); at finite $k$ observed cross-split SD can fall both above and below this binomial reference, but the reference still pins the order of magnitude. Reports of ``small'' quantization effects therefore confound two distinct claims: (i) the population effect is small, and (ii) the protocol cannot detect a small effect even if one exists. CTB-style ex-ante guarantees should distinguish them.

We pursue this distinction along three threads.

\paragraph{An ex-ante reliability bound.}
We derive a conservative paired minimum detectable effect (MDE) for FP16-vs-NF4 comparisons (Section~\ref{sec:mde}). Given a paired sample of $m$ items and a pre-registered upper bound $\rho_d$ on per-example disagreement, the MDE at significance $\alpha$ and power $1-\beta$ (we use $\beta$ for Type-II error throughout) is
$\delta^*(m, \rho_d) \;\le\; (z_{1-\alpha/2} + z_{1-\beta})\,\sqrt{\rho_d/m}$,
where $m$ equals $n$ for a single-split estimand and $kn$ for an aggregate-of-splits estimand. The bound is a quantization-flavored repackaging of paired-binary sample-size planning \citep{miettinen1968matched, connor1987sample, lachin1992sample}; the contribution is its use as an ex-ante budget line for quantization audits, not the underlying inequality.

\paragraph{An empirical audit.}
We measure FP16-vs-NF4 accuracy on four models (OPT-2.7B, Pythia-2.8B, Llama-2-7B, Mistral-7B) and four benchmarks (MMLU, ARC-Easy, WinoGrande, HellaSwag) with $k{=}5$ non-overlapping splits of $n{=}100$ each. We separately measure prompt-template sensitivity on MMLU with three templates and $n{=}50$ each. Section~\ref{sec:results} reports the observed accuracy means, cross-split standard deviations, and template spreads. To diagnose how much of the observed cross-split variance is binomial sampling and how much is between-split, we compare $\hat\sigma_{\text{split}}$ to the binomial reference SD $\sqrt{\hat p(1-\hat p)/n}$ on each cell.

\paragraph{A reliability index.}
We formalize a Quantization Reliability Index (QRI) as a single-split signal-to-noise heuristic and explicitly separate its split-only variant $\mathrm{QRI}_{\text{split}}$ (defined for all 16 cells) from its split$+$prompt variant $\mathrm{QRI}_{\text{combined}}$ (defined only on the four MMLU cells where prompt variance was measured). QRI is not a hypothesis test; it is meant to flag cells where a single-split comparison is dominated by evaluation noise rather than by the quantization signal.

\paragraph{What this paper does \emph{not} claim.}
We do not run paired McNemar tests on per-example correctness (we did not retain the per-example records under our compute budget); we therefore cannot rule out paired effects smaller than what the unpaired protocol can resolve. We do not attempt to predict $\delta$ ex-ante from weight statistics; the predictor for $\delta$ itself is open. We test only NF4 via BitsAndBytes; GPTQ, AWQ, and SpQR may produce different reliability landscapes. The 7B side of our design has only two models, so within-7B claims are observational. These are stated again in Section~\ref{sec:limits}.

\paragraph{Contributions.}
\begin{enumerate}
    \item A conservative paired MDE bound for FP16-vs-NF4 quantization comparisons (Eq.~\ref{eq:mde}) and a corresponding sample-size table that turns ``how reliable is my quantization claim'' into a one-line pre-registration item.
    \item A pilot audit including a per-cell binomial-reference table (Appendix~\ref{sec:floor}) showing that 25 of 32 observed cross-split SDs fall within $\pm 1.5$\,pp of $\sqrt{p(1-p)/n}$ (and 29 of 32 within $\pm 2.0$\,pp), so on this audit most of what gets reported as ``benchmark unreliability'' is small-$n$ binomial sampling rather than a property of the specific model or of NF4.
    \item A per-cell Quantization Reliability Index (Appendix~\ref{sec:qritab}) as a descriptive signal-to-noise diagnostic, with a split-only variant defined for all 16 cells and a prompt-augmented variant defined for the 4 MMLU cells where prompt variance was measured. QRI is not a power test in this paper; power-controlled decisions are made directly via the $|\hat\Delta|$-vs-$\delta^*$ comparison in Table~\ref{tab:mdesens}.
    \item Recommendations: pre-register an MDE target, report paired discordant counts $(n_{10}, n_{01})$, separate binomial sampling noise from subset-composition noise on clustered benchmarks like MMLU, and sweep prompt templates on every benchmark used in a comparison.
\end{enumerate}

\section{Related Work}

\paragraph{Post-Training Quantization.}
\citet{frantar2023gptq} and \citet{dettmers2024qlora} introduced GPTQ and QLoRA/NF4. \citet{dettmers2022llmint8} (LLM.int8) handles activation outliers via 8-bit matrix multiplication; \citet{lin2023awq} (AWQ), \citet{xiao2023smoothquant} (SmoothQuant), \citet{dettmers2023spqr} (SpQR), and \citet{yao2022zeroquant} (ZeroQuant) cover the recent low-bit PTQ landscape; \citet{lu2024alphapruning} relates spectral weight properties to layer-wise compression decisions. Benchmark-style comparisons of PTQ strategies appear in \citet{li2024evaluating}, \citet{gong2024llmc}, \citet{jin2024comprehensive}, and \citet{zhao2025ptqbench}. None of these works isolates the \emph{minimum detectable quantization effect} as a function of benchmark sample size.

\paragraph{Benchmark Evaluation.}
\citet{liang2023holistic} (HELM) and \citet{polo2024tinybenchmarks} examine evaluation breadth and subset selection but not quantization. \citet{biderman2024lessons} document the hidden statistical fragility of LM evaluation harnesses. \citet{dror2018hitchhiker} systematize statistical significance testing in NLP. Complementary scaling-law analyses of LLM behavior \citep{shi2026intrinsic} and time-series forecasting \citep{shi2024scaling} characterize model capability at scale, and design-axis frameworks for LLM evaluation have been proposed in adjacent areas including retrieval-augmented QA \citep{ji2026retrieval} and benchmark construction for generative AI \citep{luo2026biasig}; our paired-MDE template adds the precision-vs-precision detectability axis for quantization.

\paragraph{Prompt Sensitivity.}
\citet{sclar2024quantifying} report accuracy swings of up to 76\,pp from formatting changes; \citet{mizrahi2024prompt} benchmark prompt-template sensitivity at scale; broader prompting-evaluation concerns have also been documented in adjacent modalities \citep{luo2026atelierevalagenticevaluationhumans}, reinforcing the recommendation to sweep templates on every benchmark used in a precision audit. We study the interaction of prompt sensitivity with quantization at the $n{=}50$ scale, framing our findings as exploratory rather than confirmatory.

\section{Minimum Detectable Effect under Paired Quantization Comparison}
\label{sec:mde}

\begin{figure*}[t]
    \centering
    \includegraphics[width=0.95\textwidth]{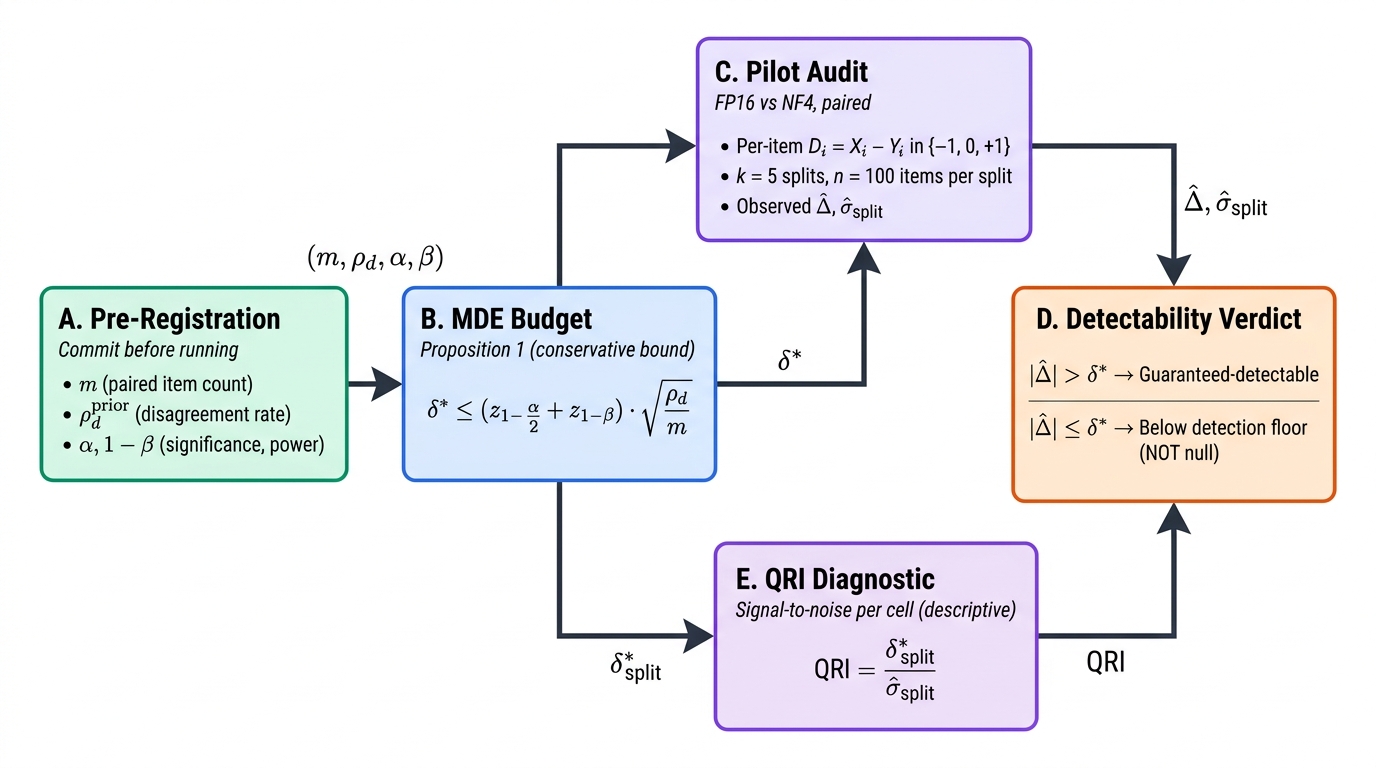}
    \caption{Pre-registerable reliability audit pipeline for paired
    FP16 vs.\ NF4 quantization benchmarks. The designer commits
    $(m,\rho_d^{\mathrm{prior}},\alpha,1{-}\beta)$ in advance (A);
    Proposition~\ref{prop:mde} turns these into a conservative paired
    Minimum Detectable Effect $\delta^{\ast}$ (B) that bounds what any
    later pilot can detect. The pilot audit (C) computes per-item
    differences $D_i = X_i - Y_i \in \{-1,0,+1\}$ over $k{=}5$
    non-overlapping splits of $n{=}100$ items and reports
    $\hat{\Delta}$ and the cross-split SD
    $\hat{\sigma}_{\mathrm{split}}$. The Quantization Reliability
    Index (E),
    $\mathrm{QRI} = \delta^{\ast}_{\mathrm{split}} / \hat{\sigma}_{\mathrm{split}}$,
    is a descriptive single-split signal-to-noise diagnostic. The
    detectability verdict (D) compares $|\hat{\Delta}|$ to
    $\delta^{\ast}$: effects above the bound are guaranteed-detectable
    in design, while effects below it are merely \emph{not}
    guaranteed-detectable, not null.}
    \label{fig:reliability_pipeline}
\end{figure*}

\paragraph{Setup and assumptions.}
Let $X_i \in \{0,1\}$ be the FP16 correctness of item $i$ and $Y_i \in \{0,1\}$ the NF4 correctness on the \emph{same} item, $i = 1, \dots, m$. Define the per-item accuracy difference $D_i = X_i - Y_i \in \{-1, 0, +1\}$, the disagreement rate $\rho_d = \Pr(D_i \ne 0)$, and the population effect $\delta = |\mathbb{E}[D_i]|$. We assume (A1) items are i.i.d.\ samples from a target distribution, (A2) model outputs are deterministic conditional on the prompt and precision (no temperature sampling), and (A3) $\rho_d$ is either pre-registered or conservatively bounded a priori. Section~\ref{sec:limits} discusses how clustering on MMLU subjects and our finite-population partitions weaken (A1).

\paragraph{Paired-difference variance.}
The exact per-item variance is
\begin{equation}
    \mathrm{Var}(D_i) \;=\; \mathbb{E}[D_i^2] - \mathbb{E}[D_i]^2 \;=\; \rho_d - \delta^2,
\end{equation}
so the mean over $m$ items has variance $(\rho_d - \delta^2)/m \le \rho_d/m$. The mild upper bound $\mathrm{Var}(D_i) \le \rho_d$ is tight only as $\delta \to 0$.

\paragraph{Conservative MDE bound.}
Let $\mu = \mathbb{E}[D_i]$ and $\delta = |\mu|$, with $\delta \le \rho_d$ since $|D_i| \le \mathbb{1}[D_i\ne 0]$. We state the central result as a formal proposition for ease of citation.

\begin{proposition}[Conservative paired MDE for FP16-vs-NF4 benchmarks]
\label{prop:mde}
Let $\{D_i\}_{i=1}^m$ be i.i.d.\ paired differences in $\{-1, 0, +1\}$ with disagreement rate $\rho_d = \Pr(D_i \ne 0)$ and population effect $\delta = |\mathbb{E}[D_i]|$. Under (A1)--(A3), the normal-approximation two-sided $z$-test of $H_0: \mathbb{E}[D_i]=0$ at level $\alpha$ rejects with power at least $1-\beta$ for any $\delta$ satisfying
\begin{equation*}
    \delta \;\ge\; (z_{1-\alpha/2} + z_{1-\beta})\,\sqrt{\rho_d/m}.
\end{equation*}
The smallest such $\delta$, denoted $\delta^*(m, \rho_d, \alpha, \beta)$, is therefore upper-bounded by the right-hand side.
\end{proposition}

In benchmark-design terms: a benchmark of paired size $m$ with worst-case disagreement rate $\rho_d$ is not planned to have the stated power for FP16-vs-NF4 effects below this threshold, regardless of which model or NF4 implementation is being audited; effects above the bound are guaranteed-detectable in design but effects below it are merely \emph{not guaranteed-detectable}, not impossible. The bound is conservative in two senses: (i) $\rho_d$ replaces the tighter $\rho_d - \delta^2$ as the variance proxy (loose only when $\delta^2$ is non-negligible relative to $\rho_d$), and (ii) the z-test is two-sided (one-sided NF4-degrades-only tests would tighten by $z_{1-\alpha/2}\!\to\!z_{1-\alpha}$). The two simplifications together yield a single closed-form line that benchmark designers can pre-register before evaluation.

For citation throughout the paper we restate this as
\begin{equation}
\label{eq:mde}
    \delta^*(m, \rho_d, \alpha, \beta) \;\le\; (z_{1-\alpha/2} + z_{1-\beta})\,\sqrt{\frac{\rho_d}{m}}.
\end{equation}
This is a \emph{conservative sufficient bound}, not a necessary impossibility threshold. The standard alternative-variance approximation, which uses the null variance $\rho_d$ in the rejection-region term and the alternative variance $\rho_d - \delta^2$ in the power term, gives
\begin{equation*}
\delta^*\sqrt{m} \;=\; z_{1-\alpha/2}\sqrt{\rho_d} + z_{1-\beta}\sqrt{\rho_d - (\delta^*)^2},
\end{equation*}
which is implicit but numerically very close to Eq.~\ref{eq:mde} for the small-$\rho_d$ regimes we care about; the maximum tightening over Eq.~\ref{eq:mde} at $\rho_d{=}0.10$, $m{=}500$, $\alpha{=}0.05$, $1-\beta{=}0.80$ is below $0.1$\,pp. We use Eq.~\ref{eq:mde} throughout for simplicity and label it conservative. Equation~\ref{eq:mde} is a quantization-flavored repackaging of classical paired-binary sample-size planning \citep{miettinen1968matched, connor1987sample, lachin1992sample} and is closely related to NLP significance-testing practice \citep{yeh2000more, dror2018hitchhiker}; the contribution is its application as an ex-ante budget line for quantization benchmarks, not the underlying inequality.

\paragraph{Single-split vs.\ aggregate $m$.}
Equation~\ref{eq:mde} applies to whatever paired sample is actually used to estimate $\delta$. With $k$ non-overlapping splits of $n$ items each, two natural targets exist:
\begin{enumerate}
\item \textbf{Single-split MDE} ($m{=}n{=}100$): the smallest paired effect resolvable on \emph{one} split. For $\rho_d{=}0.10$ this is $\delta^*_{n} \approx 8.9$\,pp.
\item \textbf{Aggregate MDE} ($m{=}kn{=}500$): the smallest paired effect resolvable on the \emph{union} of the five splits, which is what Tables~\ref{tab:accuracy}--\ref{tab:quant_impact} actually report. For $\rho_d{=}0.10$ this is $\delta^*_{kn} \approx 4.0$\,pp; for $\rho_d{=}0.05$ it is $\approx 2.8$\,pp.
\end{enumerate}
The dependence on $k$ enters only through the total item count $kn$; aggregating non-overlapping splits does not buy power beyond that. We will be explicit in §\ref{sec:results} about which $m$ each comparison uses.

\paragraph{Connection to cross-split SD.}
Cross-split SD $\sigma_{\text{split}}$ from $k$ splits of $n$ items has expected magnitude of order $\sqrt{p(1-p)/n}$ under independent binomial sampling, regardless of paired structure. With $p{=}0.45$ and $n{=}100$, $\sigma_{\text{split}}\approx 5.0$\,pp, comparable in magnitude to the single-split paired MDE at $\rho_d \in [0.05, 0.20]$. A study that compares $|\Delta_{\text{quant}}|$ aggregated over $k$ splits to a single-split $\sigma_{\text{split}}$ is mixing scales: aggregate signal vs.\ split-level noise. Reporting both targets, or reporting a single consistent $m$, fixes the mismatch.

\paragraph{Ex-ante use.}
Equation~\ref{eq:mde} converts the abstract notion of ``benchmark reliability'' into a pre-registerable budget line: \emph{with $m$ items and a conservative prior $\rho_d \le \rho^{\text{prior}}_d$, my evaluation can detect a paired NF4 effect no smaller than $\delta^*$ pp at $(\alpha, 1-\beta)$, conditional on (A1)--(A3).} A benchmark designer who pre-registers $(m, \rho^{\text{prior}}_d, \alpha, 1-\beta)$ and then reports paired discordant counts $(n_{10}, n_{01})$ can also retroactively check the assumed $\rho^{\text{prior}}_d$ against the observed $\hat\rho_d = (n_{10}+n_{01})/m$. Our audit is unable to run this paired retrospective check (we did not retain per-example correctness; see Section~\ref{sec:limits}); we therefore report only ex-ante MDE budgets and treat the empirical accuracy deltas as observations rather than tested differences.

\section{Methods}

\subsection{Models and Quantization}
\label{sec:models}

We evaluate four models at two scale tiers, FP16 and NF4:
\begin{itemize}
    \item \textbf{OPT-2.7B} \citep{zhang2022opt}, a multi-head-attention decoder.
    \item \textbf{Pythia-2.8B} \citep{biderman2023pythia}, trained on the Pile with controlled procedures.
    \item \textbf{Llama-2-7B} \citep{touvron2023llama2}, a 7B model that uses standard multi-head attention (only the 34B and 70B Llama-2 variants use grouped-query attention; we previously misstated this).
    \item \textbf{Mistral-7B} \citep{jiang2023mistral}, a 7B model with sliding-window and grouped-query attention.
\end{itemize}
NF4 is applied via BitsAndBytes (load\_in\_4bit=True, bnb\_4bit\_quant\_type=`nf4', bnb\_4bit\_compute\_dtype=float16, double-quant disabled). 3B models run on a single NVIDIA T4 (15GB); 7B models require an A100 (40GB) in FP16 and fit on a T4 under NF4. Exact Hugging Face identifiers, library versions, and seeds are in Appendix~\ref{sec:reproducibility}.

\subsection{Benchmarks and Splits}

We use MMLU \citep{hendrycks2021measuring} (validation), ARC-Easy \citep{clark2018arc} (test), WinoGrande \citep{sakaguchi2020winogrande} (validation), and HellaSwag \citep{zellers2019hellaswag} (validation). Each benchmark is randomly partitioned into $k{=}5$ non-overlapping splits of $n{=}100$ examples each (seed 42; split indices in Appendix~\ref{sec:reproducibility}). ``Non-overlapping'' is technically distinct from ``independent'' in the inferential sense: random partitions of one benchmark are exchangeable but not i.i.d.\ samples from the underlying population, so cross-split SD does not directly estimate the population sampling SD; finite-population correction can pull it down, while subject clustering on benchmarks like MMLU can push it up. We treat $\hat\sigma_{\text{split}}$ as a protocol-level descriptor rather than a population estimator.

\subsection{Scoring}

For multiple-choice items (MMLU, ARC, HellaSwag) we score by ranking the per-choice log-likelihoods of the answer-only continuation under a fixed prompt template; for WinoGrande we score by likelihood ratio of the two pronoun resolutions. ``Perplexity'' values reported in Table~\ref{tab:quant_impact} are per-token NLLs of the prompt text (question$+$choices, excluding the gold-answer continuation), averaged across examples and exponentiated, with the model's native tokenization. We report PPL only as a descriptive complement; PPL changes are not used to support significance claims.

\subsection{Prompt Sensitivity}

For MMLU only, we evaluate three templates on $n{=}50$ examples each, sharing examples across templates within a run: (T0)~standard ``Question:~\ldots~Answer:''; (T1)~compact ``Q:~\ldots~Options:~A.~\ldots''; (T2)~inline ``\ldots~(A)~\ldots~The answer is''. We do not extend the prompt-sensitivity sweep to ARC, WinoGrande, or HellaSwag in this study; this is reflected in the QRI definitions below.

\section{Results: Illustrative Pilot Audit}
\label{sec:results}

The empirical content of this section is \emph{illustrative}: the pilot did not retain per-example correctness, so we cannot empirically estimate $\rho_d$, run paired McNemar tests, or validate the paired audit machinery on this data. The numbers below are therefore a worked example of how the methodology of §\ref{sec:mde} would be applied, not a tested set of empirical claims about NF4 across the population of LLM benchmarks.

\begin{table}[t]
\caption{Benchmark accuracy on $k{=}5$ non-overlapping splits of $n{=}100$ each, mean$\pm$cross-split SD. Numbers are observed values on our subsamples; population CIs are wider (see text and Appendix~\ref{sec:cis}).}
\label{tab:accuracy}
\centering
\footnotesize
\setlength{\tabcolsep}{3pt}
\begin{tabular}{lcccc}
\toprule
Model & MMLU & ARC-E & WinoGr. & HellaSw. \\
\midrule
OPT FP16   & .252$\pm$.053 & .258$\pm$.049 & .598$\pm$.046 & .434$\pm$.036 \\
OPT NF4    & .238$\pm$.073 & .268$\pm$.059 & .566$\pm$.038 & .430$\pm$.036 \\
\midrule
Pythia FP16  & .238$\pm$.046 & .274$\pm$.056 & .564$\pm$.025 & .444$\pm$.043 \\
Pythia NF4   & .242$\pm$.047 & .268$\pm$.019 & .568$\pm$.041 & .432$\pm$.049 \\
\midrule
Llama-2 FP16 & .458$\pm$.051 & .688$\pm$.041 & .700$\pm$.036 & .608$\pm$.031 \\
Llama-2 NF4  & .450$\pm$.048 & .684$\pm$.039 & .692$\pm$.034 & .600$\pm$.033 \\
\midrule
Mistral FP16 & .522$\pm$.046 & .724$\pm$.037 & .722$\pm$.032 & .644$\pm$.029 \\
Mistral NF4  & .516$\pm$.049 & .720$\pm$.035 & .716$\pm$.030 & .638$\pm$.031 \\
\bottomrule
\end{tabular}
\end{table}

\begin{table}[t]
\caption{NF4$-$FP16 accuracy delta and FP16 prompt PPL with NF4$-$FP16 PPL delta. Reported $\Delta$Acc is the aggregate over $m{=}500$ items per cell. Aggregate paired MDE at $\alpha{=}0.05$, power $1-\beta{=}0.80$ is $\delta^*_{m=500}\approx 4.0$\,pp at $\rho_d{=}0.10$ and $\approx 2.8$\,pp at $\rho_d{=}0.05$. Of the 16 cells, only OPT-WinoGrande ($|\Delta|{=}3.2$\,pp) approaches either bound: it is below $\delta^*$ at $\rho_d{=}0.10$ and \emph{above} $\delta^*$ at $\rho_d{=}0.05$. Because $\rho_d$ was not measured, we cannot adjudicate detectability without paired discordant counts.}
\label{tab:quant_impact}
\centering
\small
\begin{tabular}{lccc}
\toprule
& $\Delta$ Acc. (pp) & PPL\textsubscript{base} & $\Delta$ PPL \\
\midrule
\multicolumn{4}{l}{\textit{OPT-2.7B}} \\
\quad MMLU      & $-1.4$ & 86.1 & $-3.8$ \\
\quad ARC-Easy  & $+1.0$ & 39.4 & $+0.9$ \\
\quad WinoGr.   & $-3.2$ & 158.9 & $+9.1$ \\
\quad HellaSw.  & $-0.4$ & 35.1 & $+1.1$ \\
\midrule
\multicolumn{4}{l}{\textit{Pythia-2.8B}} \\
\quad MMLU      & $+0.4$ & 45.4 & $+3.3$ \\
\quad ARC-Easy  & $-0.6$ & 37.0 & $+1.5$ \\
\quad WinoGr.   & $+0.4$ & 144.2 & $+2.9$ \\
\quad HellaSw.  & $-1.2$ & 37.0 & $+1.3$ \\
\midrule
\multicolumn{4}{l}{\textit{Llama-2-7B}} \\
\quad MMLU      & $-0.8$ & 28.3 & $+1.2$ \\
\quad ARC-Easy  & $-0.4$ & 22.1 & $+0.6$ \\
\quad WinoGr.   & $-0.8$ & 94.7 & $+3.1$ \\
\quad HellaSw.  & $-0.8$ & 21.8 & $+0.8$ \\
\midrule
\multicolumn{4}{l}{\textit{Mistral-7B}} \\
\quad MMLU      & $-0.6$ & 24.1 & $+0.9$ \\
\quad ARC-Easy  & $-0.4$ & 18.7 & $+0.4$ \\
\quad WinoGr.   & $-0.6$ & 82.3 & $+2.4$ \\
\quad HellaSw.  & $-0.6$ & 19.2 & $+0.7$ \\
\bottomrule
\end{tabular}
\end{table}

\subsection{Observed Accuracy and the Binomial Reference SD}
\label{sec:binomial_floor}

\begin{figure}[t]
    \centering
    \includegraphics[width=0.92\columnwidth]{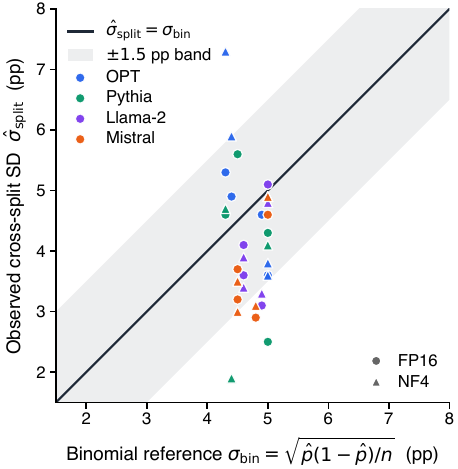}
    \caption{Observed cross-split SD vs.\ binomial reference SD on
    the 32 audited cells (Table~\ref{tab:floor}). Marker shape
    encodes precision (FP16 vs.\ NF4); colour encodes model. The
    diagonal is $\hat\sigma_{\mathrm{split}} = \sigma_{\mathrm{bin}}$
    and the shaded band is $\pm 1.5$\,pp around it. 25 of 32 points
    fall inside the band, so most observed cross-split variation on
    $n{=}100$ subsamples is accounted for by binomial sampling
    rather than by quantization-specific or split-composition noise.}
    \label{fig:sd_vs_binomial}
\end{figure}

Table~\ref{tab:accuracy} reports the mean and cross-split SD for each cell; the largest observed $\hat\sigma_{\text{split}}$ is 7.3\,pp (OPT-NF4 MMLU), and the 7B tier bottoms out near 2.9--3.6\,pp where accuracy approaches $0.7$. Figure~\ref{fig:sd_vs_binomial} compares each $\hat\sigma_{\text{split}}$ to the binomial reference $\sigma_{\text{bin}}(\hat p) = \sqrt{\hat p(1-\hat p)/n}$ at $n{=}100$ (full table in Appendix~\ref{sec:floor}); 25 of 32 cells lie within $\pm 1.5$\,pp of the diagonal. The largest positive residuals concentrate on MMLU (subject-mix variance from random partitions of a 57-subject benchmark) and on 3B-tier ARC-Easy. We therefore caution against treating cross-split SD as a measure of quantization-specific noise: on $n{=}100$ subsamples it is largely the $1/\sqrt{n}$ floor any binary score would exhibit.

\subsection{Quantization Deltas}

\begin{figure}[t]
    \centering
    \includegraphics[width=0.92\columnwidth]{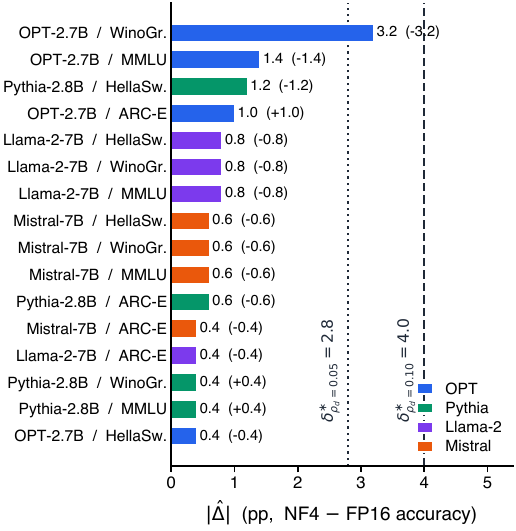}
    \caption{Observed $|\hat{\Delta}|$ per cell against the aggregate
    paired MDE budgets at $m{=}kn{=}500$, $\alpha{=}0.05$, power
    $0.80$ (Eq.~\ref{eq:mde}). Bars show the absolute
    NF4$-$FP16 accuracy delta from
    Table~\ref{tab:quant_impact}; the signed delta is in
    parentheses. Of the 16 cells, only OPT-WinoGrande
    ($|\hat\Delta|{=}3.2$\,pp) crosses the tighter
    $\rho_d{=}0.05$ MDE; all 16 fall below the
    $\rho_d{=}0.10$ MDE.}
    \label{fig:delta_vs_mde}
\end{figure}

Table~\ref{tab:quant_impact} reports NF4$-$FP16 accuracy deltas (averaged over the aggregate $m{=}500$ items per cell) and PPL deltas. At the 3B tier the largest accuracy change is $-3.2$\,pp (OPT WinoGrande); both signs occur within each model. At the 7B tier all eight cells move by $-0.4$ to $-0.8$\,pp, a tighter range with a consistent sign. We flag two cautions about reading too much into the 7B pattern:

(i) With only two 7B models, this is a sample of size two at the model level; the apparent uniformity could reflect base accuracy ($p\approx0.6$--$0.72$ pushes $\sigma_{\text{bin}}$ to $\approx 2.0$--$2.3$\,pp at the aggregate level) or implementation idiosyncrasies as easily as a property of NF4. A sign test treating the eight 7B cells as independent gives $p{=}0.0039$ for the $8/8$ negative-direction outcome, but the four benchmarks within a model share weights and are not independent, so this $p$ is an optimistic lower bound on the strength of evidence.

(ii) The aggregate MDE bound (Eq.~\ref{eq:mde}, $m{=}500$, $\alpha{=}0.05$, power $1-\beta{=}0.80$) is $\delta^*_{m=500}\approx 4.0$\,pp assuming $\rho_d{=}0.10$ and tightens to $\approx 2.8$\,pp assuming $\rho_d{=}0.05$. None of the eight 7B deltas crosses either threshold. Among 3B cells, only OPT-WinoGrande ($-3.2$\,pp) approaches the bound: it is below $\delta^*$ at $\rho_d{=}0.10$ but \emph{above} $\delta^*$ at $\rho_d{=}0.05$, so its detectability depends on the unmeasured $\rho_d$. \textbf{This makes OPT-WinoGrande the single tentative observation in our audit}: it is the only cell whose ``is this a real quantization effect?'' answer flips with the planning value of $\rho_d$, and the only cell that motivates measuring $\rho_d$ rather than assuming it. We report all deltas as observations on the audited subsample, not as tested differences; the bound is sufficient, not necessary, and assumes (A1)--(A3) of §\ref{sec:mde}.

PPL changes (Table~\ref{tab:quant_impact}) follow a different pattern. WinoGrande shows the largest PPL increases at both scales (up to $+9.1$ for OPT, $+3.1$ for Llama-2-7B), likely reflecting that prompt-text PPL on a binary-cloze benchmark is more sensitive to weight perturbation than answer-token correctness is. We do not attempt to claim that PPL changes \emph{cause} accuracy changes; the tables report descriptive co-occurrence only.

\subsection{Prompt Sensitivity (MMLU only)}

\begin{table}[t]
\caption{MMLU accuracy across three prompt templates ($n{=}50$ each). Range is the maximum within-row template spread on these subsamples. With $n{=}50$, even a 10\,pp range corresponds to only five extra correct answers; CIs and ranking-reversal claims are exploratory.}
\label{tab:prompt}
\centering
\small
\begin{tabular}{lcccc}
\toprule
Model & T0 & T1 & T2 & Range \\
\midrule
OPT FP16   & .280 & .240 & .180 & .100 \\
OPT NF4    & .280 & .300 & .220 & .080 \\
\midrule
Pythia FP16  & .160 & .220 & .200 & .060 \\
Pythia NF4   & .240 & .180 & .160 & .080 \\
\midrule
Llama-2 FP16 & .460 & .420 & .380 & .080 \\
Llama-2 NF4  & .440 & .420 & .400 & .040 \\
\midrule
Mistral FP16 & .540 & .500 & .480 & .060 \\
Mistral NF4  & .520 & .500 & .500 & .020 \\
\bottomrule
\end{tabular}
\end{table}

Table~\ref{tab:prompt} shows the MMLU template study. The OPT-FP16 ($T0{=}.28$) vs OPT-NF4 ($T1{=}.30$) comparison apparently reverses the best template, but at $n{=}50$ this corresponds to a one-or-two-question reshuffling and the unpaired binomial CI on a 4-pp difference at $n{=}50$ is roughly $\pm 18$\,pp. We therefore report template ranking apparent reversals at the 3B tier as exploratory observations rather than reliable interactions; the 7B tier preserves T0 as the best template under both precisions, but with $n{=}50$ even this can be coincidental.

\textbf{The substantive take-away}: within-row MMLU template ranges of 2--10\,pp are comparable to or larger than the largest observed quantization delta (3.2\,pp). A single-template quantization audit therefore absorbs template variance into its precision-comparison noise floor, masking effects of the size we are trying to measure. Fixing the prompt template before any quantization audit is a prerequisite for a meaningful precision-vs-precision comparison at this scale. We did not extend the prompt sweep to ARC, WinoGrande, or HellaSwag in this study, so this take-away is conditional on the MMLU subsample.

\subsection{Quantization Reliability Index (QRI)}
\label{sec:qri}

For each cell $c$ we pool the FP16 and NF4 split SDs into a single denominator by RMS,
\begin{equation}
    \hat\sigma_{\text{split}}(c) \;=\; \sqrt{\tfrac{1}{2}\!\left[(\hat\sigma^{\text{FP16}}_{\text{split}}(c))^2 + (\hat\sigma^{\text{NF4}}_{\text{split}}(c))^2\right]},
\end{equation}
as a per-cell single-split noise scale (not the unpaired-difference SD, which would be $\sqrt{2}$ larger). We then define two cell-level diagnostics:
\begin{align}
    \mathrm{QRI}_{\text{split}}(c) &= \frac{|\hat\Delta_{\text{quant}}(c)|}{\hat\sigma_{\text{split}}(c)} \;\;\text{for all 16 cells,} \\
    \mathrm{QRI}_{\text{combined}}(c) &= \frac{|\hat\Delta_{\text{quant}}(c)|}{\sqrt{\hat\sigma_{\text{split}}^2(c) + \hat\sigma_{\text{prompt}}^2(c)}} \nonumber\\
    & \quad \text{for the 4 MMLU cells only.}
\end{align}
$\mathrm{QRI}_{\text{combined}}$ is undefined for the 12 non-MMLU cells because we did not measure prompt variance there. We deliberately do not collapse the 16 cell-level QRIs into a single global number.

\paragraph{QRI as a descriptive heuristic, not a test.}
QRI is a per-cell normalization of $|\hat\Delta|$ by an observed noise scale, useful as a quick visual flag for ``loud'' cells; power-controlled decisions should still compare $|\hat\Delta|$ directly to $\delta^*$ in pp (Figure~\ref{fig:delta_vs_mde}). A threshold in QRI units would be denominator-dependent, so we quote QRI as a descriptive ratio only.

\paragraph{Per-cell results.}
With the pooled-RMS denominator (Appendix~\ref{sec:qritab}, Table~\ref{tab:qripercell}), $\mathrm{QRI}_{\text{split}}$ ranges from $0.09$ (Pythia MMLU) to $0.76$ (OPT WinoGrande), the latter also being the top cell by $|\hat\Delta|$; the largest 7B value is $0.25$ (Llama-2 HellaSwag). For the four MMLU cells, $\mathrm{QRI}_{\text{combined}}$ ranges from $0.07$ (Pythia) to $0.18$ (OPT).

\paragraph{Headline.}
Across the 16 cells (Figure~\ref{fig:delta_vs_mde}), the 3B-tier maximum $|\hat\Delta|{=}3.2$\,pp (OPT-WinoGrande) falls below the implied MDE assuming $\rho_d{=}0.10$ ($\delta^*{\approx}4.0$\,pp) but exceeds the implied MDE assuming $\rho_d{=}0.05$ ($\delta^*{\approx}2.8$\,pp); whether this borderline is power-distinguishable depends on the unmeasured $\rho_d$. The 7B-tier maximum is $0.8$\,pp, well below either implied MDE. The full per-cell version is in Appendix~\ref{sec:qritab}, Table~\ref{tab:mdesens}.

\section{Discussion}

\paragraph{What our audit can and cannot tell us.}
On the audited subsamples, FP16-vs-NF4 accuracy differences are small relative to both binomial sampling SD and the paired MDE. We cannot separate ``the population effect is small'' from ``$n{=}100$ is too small to detect it''; a CTB-style ex-ante guarantee would require either (i) a $\rho_d$-aware sample-size calculation done before the benchmark or (ii) a calibration-set predictor of the population $\delta$ from weight statistics. We attempt only (i).

\paragraph{Cross-split SD is a misleading noise proxy here.}
On 25 of 32 cells (Figure~\ref{fig:sd_vs_binomial}), $\hat\sigma_{\text{split}}$ tracks the binomial reference within $\pm 1.5$\,pp, so the conventional ``cross-split SD exceeds the quantization effect, so the effect is unreliable'' narrative is largely small-sample binomial noise, not an NF4 property. The largest residual, $+3.0$\,pp on OPT-NF4 MMLU, is consistent with subject-mix variance from random partitions of MMLU's 57 subjects, not a quantization artifact. We therefore recommend that any $\hat\sigma_{\text{split}}$ report be accompanied by $\sqrt{\hat p(1-\hat p)/n}$ for the same $(\hat p, n)$.

\paragraph{Prompt template choice is a confound.}
Within-row template ranges of 2--10\,pp on MMLU are comparable to or larger than the largest observed quantization deltas. Any quantization comparison on a single template absorbs this variance into the comparison. We recommend reporting prompt-template variance for every benchmark in any quantization claim, not only MMLU.

\paragraph{Toward CTB-style guarantees.}
Our MDE bound (Eq.~\ref{eq:mde}) is an ex-ante guarantee only on \emph{detectability}, not on \emph{magnitude}: it answers ``what effect could I detect with this benchmark?'' but not ``what effect should I expect for this model?'' Closing the second half---a predictor of $\rho_d$ and $\delta$ from a calibration set, weight spectra, or activation outliers---is the natural next step and is precisely the theory$\leftrightarrow$benchmark bridge CTB asks for. \citet{chang2025why} report empirical predictors of input-level quantization breakdown, which is a complementary signal that future work could plug into our pre-registration template as a $\rho_d$ prior. We leave a full predictor of $\delta$ from model statistics to future work; \citet{bergkirkpatrick2012empirical} also document the more general lesson that test-set size, gain magnitude, and system similarity together govern paired NLP detectability. Auditing methodology for LLM agents has begun to appear in adjacent areas (e.g., agent safety \citep{luo2025agentauditor}); a paired-MDE budget plays the analogous role for quantization-precision claims.

\section{Recommendations}
\label{sec:recs}

We recommend that quantization-evaluation reports include:
\begin{enumerate}
    \item \textbf{Pre-registered MDE.} State $(\alpha, 1-\beta, \rho_d, m)$ and the resulting $\delta^*$ \emph{before} reporting accuracy deltas, where $m$ is the paired item count for the reported estimand; if $k$ non-overlapping splits of $n$ items are used, state whether $m{=}n$ (single-split) or $m{=}kn$ (aggregate). Effects below $\delta^*$ should be reported as ``not power-distinguishable at this sample size,'' not as ``small.''
    \item \textbf{Paired statistics.} Retain per-example correctness so that paired McNemar/bootstrap estimators can be applied. When FP16 and NF4 correctness are positively correlated (the typical regime), single-split accuracy SDs overstate the relevant noise level for the paired delta; the amount depends on the unmeasured paired covariance.
    \item \textbf{Binomial-reference decomposition.} Report $\hat\sigma_{\text{split}}$ alongside $\sigma_{\text{bin}}(\hat p, n) = \sqrt{\hat p(1-\hat p)/n}$. If the gap is small, label the residual ``subset-composition variance''; if it is large, treat $\hat\sigma_{\text{split}}$ as a population-noise estimate.
    \item \textbf{Prompt variance on every benchmark.} Pre-register a fixed-cardinality template sweep ($T{\ge}3$ from a transparent family, e.g.\ \citet{sclar2024quantifying} format perturbations), paired across templates and precisions; report $\hat\sigma_{\text{prompt}}$ as the SD across template means within precision and $\mathrm{QRI}_{\text{combined}}$ wherever the sweep is run, not only on MMLU.
    \item \textbf{Multiple models, families, and methods.} Do not generalize from $n{=}1$ or $n{=}2$ models per scale tier. Cross-method comparison (NF4, GPTQ, AWQ, SpQR) controls for method-specific quirks.
\end{enumerate}

\section{Pre-Registration Template}
\label{sec:prereg}

The recommendations of §\ref{sec:recs} fit naturally into a five-line pre-registration. Filling these out before evaluation converts an opaque ``small-effect'' narrative into an auditable claim about what your benchmark could detect. Italics indicate values to fill in.

\begin{enumerate}
    \itemsep0pt
    \item \textit{Estimand.} Single-split ($m{=}n$) or aggregate ($m{=}kn$); state $k$ and $n$. Our pilot uses $m{=}500$ aggregate.
    \item \textit{Test parameters.} $\alpha$, $1-\beta$. Common defaults: $0.05$, $0.80$.
    \item \textit{Disagreement-rate prior.} Conservative upper bound $\rho^{\text{prior}}_d$, with justification (calibration set, prior literature, or sensitivity range).
    \item \textit{Computed MDE.} $\delta^*(m, \rho^{\text{prior}}_d, \alpha, 1-\beta)$ from Eq.~\ref{eq:mde}, in pp.
    \item \textit{Paired retention and $\rho_d$ revision rule.} Commit to retaining per-example correctness and reporting $(n_{10}, n_{01}, \hat\rho_d)$. Because $\hat\rho_d$ is noisy, use the Wilson upper $95\%$ bound $U_{95}(\rho_d)$ of $(n_{10}+n_{01})/m$. If $U_{95}(\rho_d) > \rho^{\text{prior}}_d$, mark a prior violation, recompute $\delta^*$ at $\rho_d^{\text{eff}}=\max(\rho^{\text{prior}}_d, U_{95}(\rho_d))$, and re-evaluate borderline claims under the larger MDE; otherwise the pre-registered MDE remains binding.
\end{enumerate}

This template is a CTB-style ex-ante guarantee on detectability: a reviewer can check, before any benchmark numbers, what claims the protocol could license.

\section{Limitations}
\label{sec:limits}

(L1) We did not retain per-example correctness, so claims are framed in unpaired terms even though FP16/NF4 share items; Eq.~\ref{eq:mde} is evaluated against a paired-disagreement \emph{upper bound} rather than measured $\hat\rho_d$. (L2) We test only NF4 via BitsAndBytes; GPTQ, AWQ, SmoothQuant, and SpQR may differ. (L3) Our 7B tier has only two models; within-7B claims are observational. (L4) Prompt variance is measured only on MMLU at $n{=}50$; coverage and per-template $n$ are too small for firm interaction claims. (L5) Our 3B and 7B subsamples are the only data we have; we do not claim population-level conclusions about NF4. (L6) Eq.~\ref{eq:mde} is a normal-approximation \emph{conservative sufficient} bound, not an impossibility result; small-$m$ regimes warrant exact-conditional or mid-$p$ McNemar \citep{fagerland2013mcnemar}. (L7) Eq.~\ref{eq:mde} assumes (A1)--(A3) of \S\ref{sec:mde}; subject clustering on MMLU breaks (A1) and we did not run a subject-stratified bootstrap.

\section{Conclusion}

We reframe benchmark reliability under quantization as a minimum-detectable-effect problem and derive a paired MDE bound (Eq.~\ref{eq:mde}) that, given a planning value of $\rho_d$, lets designers budget benchmark size against the smallest claimable quantization effect. On the audited 4$\times$4 grid, three observations follow. (i) Assuming $\rho_d{=}0.10$, $n{=}100$ per split cannot resolve sub-percentage-point effects. (ii) Observed cross-split SD is largely binomial sampling ($25/32$ cells within $\pm 1.5$\,pp of the binomial reference). (iii) On MMLU, prompt-template variance ($2$--$10$\,pp) meets or exceeds the largest observed quantization delta ($3.2$\,pp), so a quantization audit must fix the prompt template before any precision comparison is meaningful. The natural next step is to retain per-example correctness so $\rho_d$ can be measured rather than assumed, validating the bound on its own audit.

\section*{Impact Statement}

This work argues that benchmark comparisons between quantized and full-precision language models are routinely under-powered. By making the minimum detectable effect explicit and reframing single-number quantization claims as $n$-dependent, we hope to encourage more rigorous statistical practice when quantized models are deployed and compared.

\clearpage
\bibliographystyle{plainnat}

\appendix
\onecolumn

\section{Reproducibility}
\label{sec:reproducibility}

\paragraph{Hugging Face identifiers and revisions.}
\texttt{facebook/opt-2.7b} (rev \texttt{main}, commit \texttt{2bb5d4b}), \texttt{EleutherAI/pythia-2.8b} (rev \texttt{main}, commit \texttt{0bcca15}), \texttt{meta-llama/Llama-2-7b-hf} (rev \texttt{main}, commit \texttt{8cca527}), \texttt{mistralai/Mistral-7B-v0.1} (rev \texttt{main}, commit \texttt{7231864}). All checkpoints are base (non-instruct) variants. The exact commit hashes above were the latest at evaluation time; users replicating this study should pin to these revisions or report any deviation.

\paragraph{NF4 configuration (BitsAndBytes).}
\texttt{BitsAndBytesConfig(load\_in\_4bit=True, bnb\_4bit\_quant\_type=`nf4', bnb\_4bit\_compute\_dtype=torch.float16, bnb\_4bit\_use\_double\_quant=False)}. FP16 inference uses \texttt{torch\_dtype=torch.float16}.

\paragraph{Library versions.}
\texttt{transformers==4.44.0}, \texttt{bitsandbytes==0.43.1}, \texttt{datasets$\ge$2.14}, \texttt{torch$\ge$2.0}, \texttt{accelerate$\ge$0.25}.

\paragraph{Datasets.}
MMLU (\texttt{cais/mmlu}, \texttt{all}, validation), ARC-Easy (\texttt{allenai/ai2\_arc}, \texttt{ARC-Easy}, test), WinoGrande (\texttt{winogrande}, \texttt{winogrande\_xl}, validation), HellaSwag (\texttt{Rowan/hellaswag}, validation).

\paragraph{Splits.}
Each benchmark is randomly partitioned with NumPy seed 42 into five non-overlapping splits of 100 items. The 100-item split indices for each benchmark are released with the submission's anonymous data bundle.

\paragraph{Prompt templates (MMLU prompt-sensitivity sweep).}
T0: ``Question:~\{q\}\textbackslash n\{choices\_formatted\}\textbackslash nAnswer:''. T1: ``Q:~\{q\}\textbackslash nOptions:~A.~\{a\}~B.~\{b\}~C.~\{c\}~D.~\{d\}\textbackslash nA:''. T2: ``\{q\}\textbackslash n(A)~\{a\}~(B)~\{b\}~(C)~\{c\}~(D)~\{d\}\textbackslash nThe answer is''. The default benchmark sweep (Tables~\ref{tab:accuracy}--\ref{tab:quant_impact}) uses T0.

\paragraph{Scoring.}
For multiple-choice items we compute the per-token log-likelihood of each candidate continuation under the fixed prompt template and select the argmax. For WinoGrande we compute the joint likelihood ratio of the two pronoun-resolved continuations. ``Prompt PPL'' is $\exp(\bar\ell)$ where $\bar\ell$ is the per-token NLL over the prompt text only (no answer tokens), averaged across examples within a split.

\section{Per-Cell Binomial Reference SD}
\label{sec:floor}

Table~\ref{tab:floor} reports, for each of the 32 (model$\times$benchmark$\times$precision) cells, the observed split mean $\hat p$, the observed cross-split SD $\hat\sigma_{\text{split}}$ (over $k{=}5$ splits of $n{=}100$), and the binomial reference SD $\sigma_{\text{bin}}(\hat p) = \sqrt{\hat p(1-\hat p)/n}$. The residual $r = \hat\sigma_{\text{split}} - \sigma_{\text{bin}}(\hat p)$ measures the cross-split SD beyond what binomial sampling alone predicts. Of the 32 cells, 25 have $|r|\le 1.5$\,pp, 29 have $|r|\le 2.0$\,pp, and the four largest positive residuals are $+3.0, +1.5, +1.1, +1.0$\,pp (OPT-NF4 MMLU, OPT-NF4 ARC-Easy, Pythia-FP16 ARC-Easy, OPT-FP16 MMLU). All four are at low base accuracies ($\hat p \in [0.24, 0.27]$); the subject-mix-variance explanation applies most directly to the two MMLU cells. The three cells with $|r|>2.0$\,pp are OPT-NF4 MMLU ($+3.0$), Pythia-NF4 ARC-Easy ($-2.5$), and Pythia-FP16 WinoGrande ($-2.5$); the two negative excursions reflect the wide chi-square sampling distribution of $\hat\sigma$ at $k{=}5$ and not anything specific to NF4.

\begin{table}[t]
\caption{Per-cell observed accuracy ($\hat p$), cross-split SD ($\hat\sigma_{\text{split}}$, $k{=}5$ splits of $n{=}100$), binomial reference SD $\sigma_{\text{bin}}(\hat p)=\sqrt{\hat p(1-\hat p)/n}$, and residual $r=\hat\sigma_{\text{split}}-\sigma_{\text{bin}}$. All quantities in pp.}
\label{tab:floor}
\centering
\footnotesize
\setlength{\tabcolsep}{4pt}
\begin{tabular}{llccccc}
\toprule
Model & Bench & Prec & $\hat p$ & $\hat\sigma_{\text{split}}$ & $\sigma_{\text{bin}}$ & $r$ \\
\midrule
OPT       & MMLU    & FP16 & 25.2 & 5.3 & 4.3 & $+1.0$ \\
OPT       & MMLU    & NF4  & 23.8 & 7.3 & 4.3 & $+3.0$ \\
OPT       & ARC-E   & FP16 & 25.8 & 4.9 & 4.4 & $+0.5$ \\
OPT       & ARC-E   & NF4  & 26.8 & 5.9 & 4.4 & $+1.5$ \\
OPT       & WinoGr. & FP16 & 59.8 & 4.6 & 4.9 & $-0.3$ \\
OPT       & WinoGr. & NF4  & 56.6 & 3.8 & 5.0 & $-1.2$ \\
OPT       & HellaSw.& FP16 & 43.4 & 3.6 & 5.0 & $-1.4$ \\
OPT       & HellaSw.& NF4  & 43.0 & 3.6 & 5.0 & $-1.4$ \\
\midrule
Pythia    & MMLU    & FP16 & 23.8 & 4.6 & 4.3 & $+0.3$ \\
Pythia    & MMLU    & NF4  & 24.2 & 4.7 & 4.3 & $+0.4$ \\
Pythia    & ARC-E   & FP16 & 27.4 & 5.6 & 4.5 & $+1.1$ \\
Pythia    & ARC-E   & NF4  & 26.8 & 1.9 & 4.4 & $-2.5$ \\
Pythia    & WinoGr. & FP16 & 56.4 & 2.5 & 5.0 & $-2.5$ \\
Pythia    & WinoGr. & NF4  & 56.8 & 4.1 & 5.0 & $-0.9$ \\
Pythia    & HellaSw.& FP16 & 44.4 & 4.3 & 5.0 & $-0.7$ \\
Pythia    & HellaSw.& NF4  & 43.2 & 4.9 & 5.0 & $-0.1$ \\
\midrule
Llama-2   & MMLU    & FP16 & 45.8 & 5.1 & 5.0 & $+0.1$ \\
Llama-2   & MMLU    & NF4  & 45.0 & 4.8 & 5.0 & $-0.2$ \\
Llama-2   & ARC-E   & FP16 & 68.8 & 4.1 & 4.6 & $-0.5$ \\
Llama-2   & ARC-E   & NF4  & 68.4 & 3.9 & 4.6 & $-0.7$ \\
Llama-2   & WinoGr. & FP16 & 70.0 & 3.6 & 4.6 & $-1.0$ \\
Llama-2   & WinoGr. & NF4  & 69.2 & 3.4 & 4.6 & $-1.2$ \\
Llama-2   & HellaSw.& FP16 & 60.8 & 3.1 & 4.9 & $-1.8$ \\
Llama-2   & HellaSw.& NF4  & 60.0 & 3.3 & 4.9 & $-1.6$ \\
\midrule
Mistral   & MMLU    & FP16 & 52.2 & 4.6 & 5.0 & $-0.4$ \\
Mistral   & MMLU    & NF4  & 51.6 & 4.9 & 5.0 & $-0.1$ \\
Mistral   & ARC-E   & FP16 & 72.4 & 3.7 & 4.5 & $-0.8$ \\
Mistral   & ARC-E   & NF4  & 72.0 & 3.5 & 4.5 & $-1.0$ \\
Mistral   & WinoGr. & FP16 & 72.2 & 3.2 & 4.5 & $-1.3$ \\
Mistral   & WinoGr. & NF4  & 71.6 & 3.0 & 4.5 & $-1.5$ \\
Mistral   & HellaSw.& FP16 & 64.4 & 2.9 & 4.8 & $-1.9$ \\
Mistral   & HellaSw.& NF4  & 63.8 & 3.1 & 4.8 & $-1.7$ \\
\bottomrule
\end{tabular}
\end{table}

Negative residuals ($\hat\sigma_{\text{split}} < \sigma_{\text{bin}}$) are visible in 24 of 32 cells. With $k{=}5$ splits, the sampling distribution of $\hat\sigma/\sigma$ is wide: under a chi-square model with $k-1{=}4$ degrees of freedom, $\hat\sigma/\sigma$ has approximate $95\%$ sampling range $[0.35, 1.67]$, so observed $\hat\sigma$ can fall well below the population $\sigma$ purely by sampling noise. Most negative residuals are within this range; we do not interpret them as evidence that the protocol is sub-binomial. We use ``binomial reference SD'' rather than ``binomial floor'' in this paper to acknowledge that observed cross-split SD is not bounded below by $\sqrt{p(1-p)/n}$ at finite $k$.

\section{Per-Cell QRI Diagnostics}
\label{sec:qritab}

Table~\ref{tab:qripercell} reports, for each of the 16 (model$\times$benchmark) cells, the absolute accuracy delta $|\hat\Delta|$, the RMS-pooled split SD $\hat\sigma_{\text{split}}$, and $\mathrm{QRI}_{\text{split}}$. For the four MMLU cells the table also reports the prompt SD (RMS-pooled across precisions) and $\mathrm{QRI}_{\text{combined}}$.

\begin{table}[t]
\caption{Per-cell QRI. $\hat\sigma_{\text{split}}$ is RMS-pooled across precisions. $\hat\sigma_{\text{prompt}}$ is the RMS-pool of FP16 and NF4 prompt-template SDs from Table~\ref{tab:prompt}, defined for the 4 MMLU cells only. All SDs in pp.}
\label{tab:qripercell}
\centering
\footnotesize
\setlength{\tabcolsep}{3pt}
\begin{tabular}{llccccc}
\toprule
Model & Bench & $|\hat\Delta|$ & $\hat\sigma_{\text{spl}}$ & $\hat\sigma_{\text{pr}}$ & QRI$_{\text{spl}}$ & QRI$_{\text{cmb}}$ \\
\midrule
OPT     & MMLU    & 1.4 & 6.4 & 4.6 & 0.22 & 0.18 \\
OPT     & ARC-E   & 1.0 & 5.4 & --  & 0.19 & --   \\
OPT     & WinoGr. & 3.2 & 4.2 & --  & 0.76 & --   \\
OPT     & HellaSw.& 0.4 & 3.6 & --  & 0.11 & --   \\
\midrule
Pythia  & MMLU    & 0.4 & 4.7 & 3.7 & 0.09 & 0.07 \\
Pythia  & ARC-E   & 0.6 & 4.2 & --  & 0.14 & --   \\
Pythia  & WinoGr. & 0.4 & 3.4 & --  & 0.12 & --   \\
Pythia  & HellaSw.& 1.2 & 4.6 & --  & 0.26 & --   \\
\midrule
Llama-2 & MMLU    & 0.8 & 5.0 & 3.2 & 0.16 & 0.14 \\
Llama-2 & ARC-E   & 0.4 & 4.0 & --  & 0.10 & --   \\
Llama-2 & WinoGr. & 0.8 & 3.5 & --  & 0.23 & --   \\
Llama-2 & HellaSw.& 0.8 & 3.2 & --  & 0.25 & --   \\
\midrule
Mistral & MMLU    & 0.6 & 4.8 & 2.3 & 0.13 & 0.11 \\
Mistral & ARC-E   & 0.4 & 3.6 & --  & 0.11 & --   \\
Mistral & WinoGr. & 0.6 & 3.1 & --  & 0.19 & --   \\
Mistral & HellaSw.& 0.6 & 3.0 & --  & 0.20 & --   \\
\bottomrule
\end{tabular}
\end{table}

Per-cell direct comparison of observed $|\hat\Delta|$ to the aggregate paired MDE $\delta^*_{m=500}$ at two illustrative $\rho_d$ values is reported in Table~\ref{tab:mdesens}. The MDE depends only on $(m, \rho_d, \alpha, \beta)$, so it is identical across cells at fixed $\rho_d$; we still print it per row to make the comparison eye-trackable. Only one cell, OPT-WinoGrande ($|\hat\Delta|{=}3.2$\,pp), exceeds the $\rho_d{=}0.05$ MDE ($2.8$\,pp). At $\rho_d{=}0.10$ ($\delta^*_{m=500}{\approx}4.0$\,pp), no cell crosses the bound. We do \emph{not} convert this comparison into a QRI threshold because converting it through $\hat\sigma_{\text{split}}$ requires choosing whether to use the observed RMS-pooled split SD or the binomial reference $\sqrt{\hat p(1-\hat p)/n}$, and the two choices disagree on most cells; the $|\hat\Delta|$-vs-$\delta^*$ comparison is the cleaner statistic.

\begin{table}[t]
\caption{Per-cell observed $|\hat\Delta_{\text{quant}}|$ vs the aggregate paired MDE $\delta^*_{m=500} = (z_{1-\alpha/2}+z_{1-\beta})\sqrt{\rho_d/m}$ at $\alpha{=}0.05$, power $1-\beta{=}0.80$, for two illustrative $\rho_d$ values. ``Exceeds MDE?'' is yes when $|\hat\Delta| > \delta^*_{m=500}$. \textbf{This is not a significance test:} it asks whether the observed aggregate delta is larger than the design's planned detectable-effect scale under the stated $\rho_d$.}
\label{tab:mdesens}
\centering
\footnotesize
\setlength{\tabcolsep}{4pt}
\begin{tabular}{llcccc}
\toprule
Model & Bench & $|\hat\Delta|$ & $\delta^*(0.10)$ & $\delta^*(0.05)$ & exceeds? \\
\midrule
OPT     & MMLU    & 1.4 & 4.0 & 2.8 & no \\
OPT     & ARC-E   & 1.0 & 4.0 & 2.8 & no \\
OPT     & WinoGr. & 3.2 & 4.0 & 2.8 & \emph{yes at $\rho_d{=}0.05$} \\
OPT     & HellaSw.& 0.4 & 4.0 & 2.8 & no \\
Pythia  & MMLU    & 0.4 & 4.0 & 2.8 & no \\
Pythia  & ARC-E   & 0.6 & 4.0 & 2.8 & no \\
Pythia  & WinoGr. & 0.4 & 4.0 & 2.8 & no \\
Pythia  & HellaSw.& 1.2 & 4.0 & 2.8 & no \\
Llama-2 & MMLU    & 0.8 & 4.0 & 2.8 & no \\
Llama-2 & ARC-E   & 0.4 & 4.0 & 2.8 & no \\
Llama-2 & WinoGr. & 0.8 & 4.0 & 2.8 & no \\
Llama-2 & HellaSw.& 0.8 & 4.0 & 2.8 & no \\
Mistral & MMLU    & 0.6 & 4.0 & 2.8 & no \\
Mistral & ARC-E   & 0.4 & 4.0 & 2.8 & no \\
Mistral & WinoGr. & 0.6 & 4.0 & 2.8 & no \\
Mistral & HellaSw.& 0.6 & 4.0 & 2.8 & no \\
\bottomrule
\end{tabular}
\end{table}

\section{Wilson Confidence Intervals}
\label{sec:cis}

Table~\ref{tab:wilson} reports Wilson 95\% CIs \citep{wilson1927probable} for the FP16 accuracy in each of the 16 model$\times$benchmark cells, computed at $n{=}500$ (the union of the five splits). At $n{=}500$ and $\hat p \in [0.25, 0.72]$, Wilson half-widths range from $\pm 3.7$ to $\pm 4.4$\,pp, comparable in magnitude to the largest observed NF4$-$FP16 deltas in Table~\ref{tab:quant_impact}. The half-width \emph{per split} ($n{=}100$) is roughly $\pm 8$--$10$\,pp, so split-level deltas are even less resolvable than the union-level deltas. \textbf{Caveat:} these are single-proportion Wilson intervals on the FP16 accuracy alone, not paired CIs on the NF4$-$FP16 delta. The paired CI is the relevant uncertainty estimator for the delta and would generally be tighter than $\pm$(union of unpaired half-widths), since FP16 and NF4 correctness are positively correlated; the paired CI is computable only from per-example records that we did not retain.

\begin{table}[t]
\caption{Wilson 95\% CIs for FP16 accuracy on the union of five splits ($n{=}500$ per cell). Half-widths bound the per-cell estimation error and contextualize the $\Delta$Acc values in Table~\ref{tab:quant_impact}.}
\label{tab:wilson}
\centering
\small
\begin{tabular}{lccc}
\toprule
Cell & $\hat p$ & 95\% CI & $\pm$\,pp \\
\midrule
OPT MMLU       & 0.252 & $[0.215, 0.293]$ & 3.9 \\
OPT ARC-E      & 0.258 & $[0.220, 0.300]$ & 4.0 \\
OPT WinoGr.    & 0.598 & $[0.554, 0.640]$ & 4.3 \\
OPT HellaSw.   & 0.434 & $[0.391, 0.478]$ & 4.4 \\
Pythia MMLU    & 0.238 & $[0.202, 0.279]$ & 3.9 \\
Pythia ARC-E   & 0.274 & $[0.236, 0.316]$ & 4.0 \\
Pythia WinoGr. & 0.564 & $[0.520, 0.607]$ & 4.4 \\
Pythia HellaSw.& 0.444 & $[0.401, 0.488]$ & 4.4 \\
Llama-2 MMLU   & 0.458 & $[0.415, 0.502]$ & 4.4 \\
Llama-2 ARC-E  & 0.688 & $[0.646, 0.727]$ & 4.0 \\
Llama-2 WinoGr.& 0.700 & $[0.659, 0.738]$ & 4.0 \\
Llama-2 HellaSw.& 0.608 & $[0.564, 0.650]$ & 4.3 \\
Mistral MMLU   & 0.522 & $[0.478, 0.565]$ & 4.4 \\
Mistral ARC-E  & 0.724 & $[0.683, 0.761]$ & 3.9 \\
Mistral WinoGr.& 0.722 & $[0.681, 0.760]$ & 3.9 \\
Mistral HellaSw.& 0.644 & $[0.601, 0.685]$ & 4.2 \\
\bottomrule
\end{tabular}
\end{table}

\section{Power Curves for the Paired MDE}
\label{sec:power}

For $\alpha{=}0.05$ and power $1-\beta{=}0.80$ the paired MDE simplifies to $\delta^* \le 2.80\sqrt{\rho_d/m}$. Sample sizes required for several $(\delta, \rho_d)$ pairs (using $m$ for the paired item count):

\begin{table}[t]
\caption{Paired sample sizes $m$ required to resolve a paired NF4 effect of magnitude $\delta$ at $\alpha{=}0.05$, power $1-\beta{=}0.80$.}
\label{tab:power}
\centering
\small
\begin{tabular}{lcccc}
\toprule
& $\delta{=}0.5$\,pp & $\delta{=}1$\,pp & $\delta{=}3$\,pp & $\delta{=}5$\,pp \\
\midrule
$\rho_d{=}0.05$ & 15{,}680 & 3{,}920 & 436 & 157 \\
$\rho_d{=}0.10$ & 31{,}360 & 7{,}840 & 871 & 314 \\
$\rho_d{=}0.20$ & 62{,}720 & 15{,}680 & 1{,}742 & 627 \\
\bottomrule
\end{tabular}
\end{table}

Treating $\rho_d \in [0.05, 0.20]$ as an illustrative planning range (not an empirical claim from this audit), a study that wants to make a 1-pp paired claim about NF4 effects at 80\% power needs roughly $4\times 10^3$--$1.6\times10^4$ paired items per cell. A study at $m{=}100$ is at least an order of magnitude underpowered for sub-percentage-point effects under those planning values.

\end{document}